\documentclass[letterpaper,10pt,conference]{ieeeconf}
\IEEEoverridecommandlockouts
\usepackage{amsmath,amssymb,amsfonts}
\usepackage{graphicx}
\usepackage{subcaption}
\usepackage{cite}
\usepackage{siunitx}
\usepackage{bm}
\usepackage{booktabs}
\usepackage{multirow}
\usepackage{url}

\usepackage[T1]{fontenc}

\title{\LARGE \bf An Augmented Reality Brain-Robot Interface\\for Generalist Robot Arm Manipulation*}

\author{Shangkai Zhang$^{1}$, Rousslan Fernand Julien Dossa$^{1}$, Luca Nunziante$^{1}$, \\ Marina Di Vincenzo$^{1}$, and Kai Arulkumaran$^{1}$%
\thanks{*This work was supported by JST under Moonshot R\&D Grant Number JPMJMS2012.}%
\thanks{$^{1}$All authors were with Araya Inc., Chiyoda, Tokyo, Japan at the time of this work. 
        Email: {\tt\small zhangshangkai@g.ecc.u-tokyo.ac.jp}, 
        {\tt\small \{dossa, luca\_nunziante, marina\_di\_vincenzo, kai\_arulkumaran\}@araya.org}}%
}

\begin{document}

\maketitle
\thispagestyle{empty}
\pagestyle{empty}

\begin{abstract}
The integration of augmented reality (AR) and EEG-based brain-computer interfaces (BCIs) offers a promising path for enabling intuitive control of robots for assistive purposes. However, existing AR brain-robot interface (BRI) systems are often constrained to task-specific structures, limiting their utility in real-world environments. We present an AR BRI designed for generalist robot arm manipulation that combines gaze-based object selection with motor imagery action control. Our system uses eye-tracking for intuitive object targeting and context-aware visual overlays (``Place'' and ``Use'') to guide the user through tasks within a shared autonomy framework. We evaluated the interface through a feasibility study with 18 healthy participants performing three multi-step activities of daily living: drinking, using a drawer, and operating an oven. Our results demonstrate that this interaction paradigm enables effective sequential task execution and high user engagement, achieving a ``Good'' usability rating (SUS > 70). These findings support the feasibility of the proposed interaction paradigm for complex BCI-driven robotic assistance, and motivate future evaluation with the intended target population. Project website: \url{https://ar-bri-manip.github.io/}.
\end{abstract}

\section{INTRODUCTION}
Research in assistive robots to support individuals with physical disabilities in performing activities of daily living (ADLs) has received considerable attention in recent years. However, despite progress in technical capabilities, recent evaluations indicate that existing systems often lack the flexibility and functional breadth required for effective deployment in real-world contexts~\cite{sorensen2025humanoid}. In particular, research in physically assistive robotics has frequently progressed within isolated application domains, such as feeding or mobility assistance, limiting the development of generalist systems capable of handling sequential and non-predefined tasks~\cite{nanavati2024physically}.

%\edit{Thus, our work addresses this limitation through a multimodal interface that combines brain-based control with augmented reality to support both user input and system feedback. Multimodal interfaces combine complementary modalities to distribute interaction demands across different forms of input and feedback. Prior work has shown that multimodal communication becomes increasingly relevant under higher cognitive load~\cite{oviatt2004we}, while the integration of multiple modalities in robot teleoperation can improve task performance and reduce perceived workload~\cite{triantafyllidis2020study}.}

Recent developments point towards this changing in the near future. While early assistive robotics relied on direct teleoperation and motion primitives for complex actions~\cite{quere2020shared}, the field of robotics in general has shifted toward scalable learning approaches, with growing dataset size, diversity, and model capacity~\cite{shafiullah2023bringing}. The predominant approach now is vision-language-action (VLA) models, that fine-tune pretrained vision-language models (VLMs) to generate robot actions directly~\cite{zitkovich2023rt,o2024open,kim2024openvla,black2024pi_0,intelligence2025pi}. By leveraging the knowledge within foundation models, they provide better multi-task performance than most end-to-end engineered approaches. Yet their practical utility for assistive robotics remains limited by the interface---in particular, for people who are severely paralysed and unable to speak. In this work, we address this through the development of a multimodal brain-robot interface (BRI), combining augmented reality (AR) and electroencephalography (EEG) for user input and system feedback. Together these allow for intuitive control, with AR for visual feedback and gaze-based object selection, and EEG-based motor imagery (MI) for action selection. % By using complementary modalities, we distribute interaction demands, which can improve task performance and reduce perceived workload~\cite{oviatt2004we,triantafyllidis2020study}.

\begin{figure}[tp]
\centering
\includegraphics[width=0.75\linewidth]{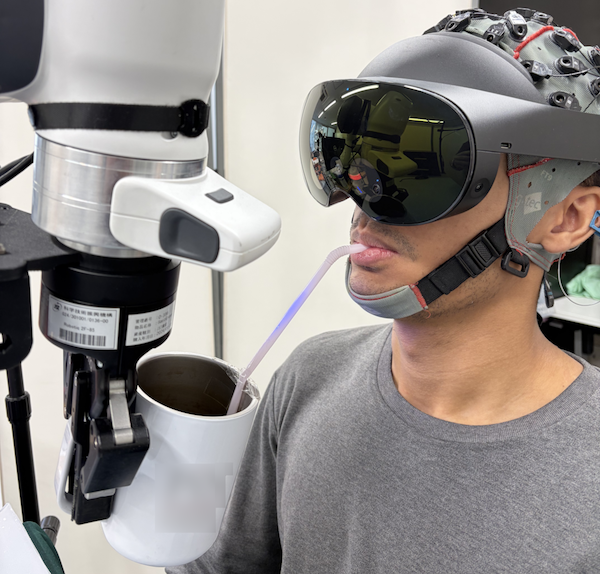}
\caption{User taking a drink using our system. After selecting the mug with eye-tracking and selecting the action with EEG, the robot brings the mug close to the user's face.}
\label{fig:drinking}
\end{figure}

Historically, many assistive robot systems have relied on screen-based or indirect control paradigms, splitting user attention between a display and the physical environment~\cite{fischer2024scoping}, with some users even preferring non-visual interfaces as a consequence~\cite{arulkumaran2024comparison}. The use of external displays increases cognitive load and reduces intuitiveness, which is even more of an issue when continuous situational awareness is required. Nowadays, AR systems have become feasible, offering a complementary interface paradigm by displaying digital information overlaid on the physical environment.

Prior work has extensively explored the use of AR for human–robot interaction and collaboration, particularly to facilitate shared understanding, visualize robot intent, and provide spatially aligned guidance cues \cite{chang2024survey}. However, these approaches primarily focus on visualization and communication, and typically rely on conventional input modalities, limiting the direct integration of user intent within the interaction loop. More recent efforts have also investigated AR as an interface for robot control, including integrations with BRIs~\cite{dillen2024shared, wang2025eeg}. However, while these approaches demonstrate the potential of combining AR with BRIs, they remain largely limited to structured manipulation scenarios and relatively simple tasks, and do not address flexible, multi-step interaction in more general real-world settings.

%\edit{In this work, we propose a multimodal, multi-device AR brain--robot interface (BRI) for generalist robot arm manipulation in ADLs. Rather than treating AR only as a visualization layer, our system integrates an AR headset with eye tracking and an external EEG interface within a unified interaction loop. Gaze enables spatial object selection, EEG motor imagery provides high-level action commands, and AR presents visual feedback aligned with the physical environment. By assigning object selection, action specification, and system feedback to complementary modalities and devices, the interface supports direct communication of user intent while preserving awareness of the robot and its surroundings.}

Going beyond these prior works, we combine our multimodal AR BRI with a generalist robot policy~\cite{intelligence2025pi} to allow users to perform ADL-inspired tasks involving multiple steps and manipulation actions. Our main contributions are: (1) an AR interface integrating spatial object selection with real-world robot execution; (2) a control framework combining eye-tracking and EEG-based MI within a shared autonomy paradigm; and (3) a feasibility study with healthy participants evaluating sequential multi-step manipulation tasks representative of ADLs. Users achieved nearly perfect success rates, were engaged, and reported low workload demands and high usability. These results support our design choices and motivate future studies with a target population that would benefit from more general-purpose assistive robotics.

\section{BACKGROUND}

\subsection{Generalist Robot Policies}
The trend of foundation models in AI has carried over into robotics within the last few years~\cite{zitkovich2023rt,o2024open,firoozi2025foundation,kawaharazuka2025vision}. By using the broad representations of pre-trained VLMs, VLAs can even generalise across unseen scenarios, objects, language instructions, and semantic concepts~\cite{kim2024openvla}.

VLAs have been applied across tasks ranging from tabletop pick-and-place~\cite{kim2024openvla,shukor2025smolvla} to complex manipulation like laundry folding~\cite{black2024pi_0,intelligence2025pi}, long-horizon recipe following~\cite{torne2026mem}, and precise skills like screw installation~\cite{xu2026rlt}. Despite these advances, however, their application to physical human-robot interaction and assistive robotics remains limited. To address this gap, we fine-tune the $\pi_{0.5}$ VLA~\cite{intelligence2025pi} on in-house demonstrations of tasks such as opening and closing an oven, and bringing a mug towards a person, and deploy this model as part of our AR BRI system.

\subsection{Shared Autonomy in Assistive Robotics}
Control of robots can range from fully manual (teleoperation) to fully autonomous. Shared autonomy sits in the middle, combining human high-level intent with autonomous policies that handle low-level motion control and environmental adaptation, making it a good fit for assistive robotics~\cite{quere2020shared}. This is especially relevant for physically assistive robots, where users may need to remain in control of task goals while avoiding continuous low-level operation during activities of daily living~\cite{nanavati2024physically}.
Indeed, recent trends in shared autonomy highlight assistive robotics and general manipulation as the two most prevalent application domains~\cite{hagenow2025shared}.

Given the possible diverse instantiation of human-robot environments, implementations of shared autonomy vary significantly. Recent perspectives on human--robot collaboration also emphasize the importance of intuitive communication, adaptive task planning, dynamic role allocation, and feedback between the human and robot~\cite{lagomarsino2025intuitive}.
In contrast with approaches that blend control coming from the robot and the human~\cite{atan2025probabilistic}, in our approach the human and the robot take turns to complete the overall task.
Here, the human only provides high level intent, rather than low level control like in other turn-taking frameworks~\cite{bustamante2022cats}. In this way, the user specifies what should be done, while the robot determines how to execute the selected manipulation policy.

\subsection{BRIs and AR for Robot Control}

\begin{figure*}[tbh!]
\centering
    \includegraphics[width=\linewidth]{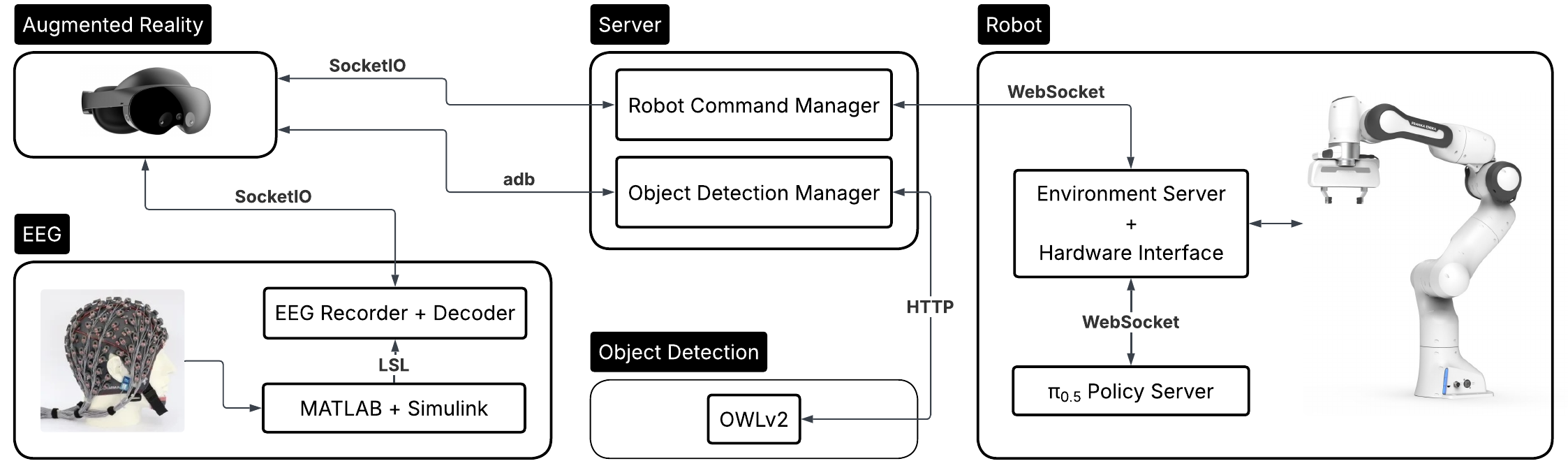}
    \caption{Modular system architecture for our multimodal AR BRI.}
    \label{fig:system_overview}
\end{figure*}

Non-invasive BRIs have been widely explored for assistive robotics, particularly for individuals with severe motor impairments \cite{millan2010combining, jamil2021noninvasive}.
Their integration with AR has emerged as a promising approach to enhance human–robot interaction through intuitive visual feedback \cite{lahtinen2024brain}.

Most existing AR BRI studies rely on stimulus-driven paradigms, particularly steady-state visually evoked potentials (SSVEP), which naturally integrate with visual overlays for target selection.
SSVEP-based approaches have been successfully applied in AR settings~\cite{liu2019ssvep}, with, e.g., robotic arm manipulation~\cite{chen2020combination} and navigation tasks~\cite{zhang2022humanoid, wang2024development}.

More broadly, MI remains one of the most extensively studied BRI paradigms due to its alignment with natural motor planning and execution.
By decoding sensorimotor rhythm modulations from EEG, MI-based systems enable users to trigger discrete commands through imagined movements~\cite{pfurtscheller2001motor, lotte2018review}.
Unlike stimulus-driven approaches, MI offers a more direct, internally-generated form of control, though it typically requires user training and careful interface design to reduce cognitive burden~\cite{jeunet2016advances}.
Consequently, its integration with AR remains less explored compared to SSVEP-based approaches, despite its potential for more natural and user-driven interaction.

Recent work has begun exploring MI-based BRI integration with immersive technologies for assistive robotics.
Dillen et al. \cite{dillen2024shared} proposed a shared control framework combining MI, eye-tracking, and AR, showing that spatial visual feedback can improve system transparency and user understanding of robot actions. 
Wang et al. \cite{wang2025eeg} also introduced an EEG-driven AR system for robotic grasping, using visual overlays to support object selection and interaction with the environment.

These prior works were limited to pick-and-place tasks with cubes. Through the use of generalist robot policies, in our work we demonstrate an AR BRI system that can be used to accomplish multi-step tasks involving other forms of manipulation, such as opening and closing different types of containers, as well as picking and placing different objects.

\section{AUGMENTED REALITY BRAIN-ROBOT INTERFACE FOR MANIPULATION}
\label{sec:mrbri}
To address the gaps identified in contemporary AR BRI systems, in this work we present a novel pipeline that enables intuitive robotic arm manipulation for a range of different tasks. Our system integrates gaze-based object selection, MI-based action selection, error recovery mechanisms, and a shared autonomy controller, with visual and textual overlays providing real-time feedback within the user's physical workspace. The following subsections detail each component of our system.

\subsection{System Architecture}
\label{sec:mrbri:system_architecture}
Our architecture is composed of five modules, as illustrated in Fig.~\ref{fig:system_overview}.
The AR module runs as an app on a Meta Quest Pro headset\footnote{https://www.meta.com/quest/quest-pro/}, and communicates with other modules via SocketIO\footnote{https://github.com/miguelgrinberg/python-socketio; https://github.com/getnamo/SocketIOClient-Unreal}.
The app was developed in Unreal Engine 5.4 to leverage passthrough and eye-tracking for hands-free interaction.
The EEG module receives 22-channel EEG data from a g.tec g.GAMMAcap2\footnote{https://www.gtec.at/product/g-scarabeo-eeg-electrodes} via a Lab Streaming Layer (LSL)~\cite{kothe2025lsl} stream. 
For deployment, we collect user data to train a scikit-learn~\cite{pedregosa2011scikitlearn} classifier, then deploy this in the EEG module to decode user intent.
The object detection module employs OWLv2~\cite{minderer2024scaling} to identify fixated objects from periodic headset screenshots. The robot module interfaces with a Franka Emika Panda arm\footnote{https://franka.de/ (model discontinued)} via RoboHive \cite{vikash2020robohive}, executing actions through a fine-tuned $\pi_{0.5}$ policy~\cite{intelligence2025pi}. Finally, an asynchronous Python-based server, extended from prior work~\cite{yoshida2025m4bench, douglas2025levels}, coordinates all modules.

While all modules can run on a single machine, computation overload and latency can cause poor performance.
Our modular approach allows the distribution of modules across separate machines, minimising overload and achieving near-real-time performance.

\begin{figure}
\centering
    \begin{subfigure}{0.235\textwidth}
        \centering
        \includegraphics[width=\linewidth]{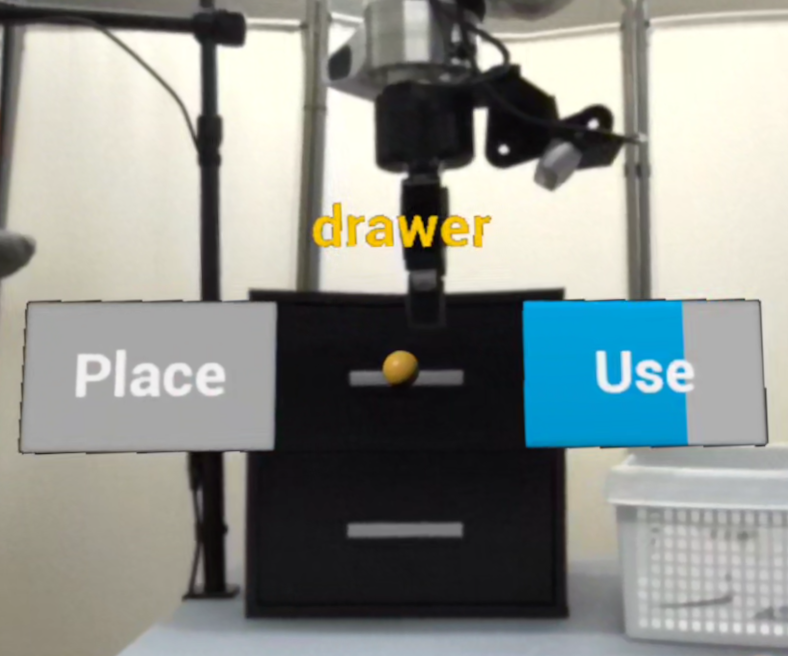}
        \label{fig:mi}
    \end{subfigure}
    %\hspace{0.035\textwidth}
    \begin{subfigure}{0.235\textwidth}
        \centering
        \includegraphics[width=\linewidth]{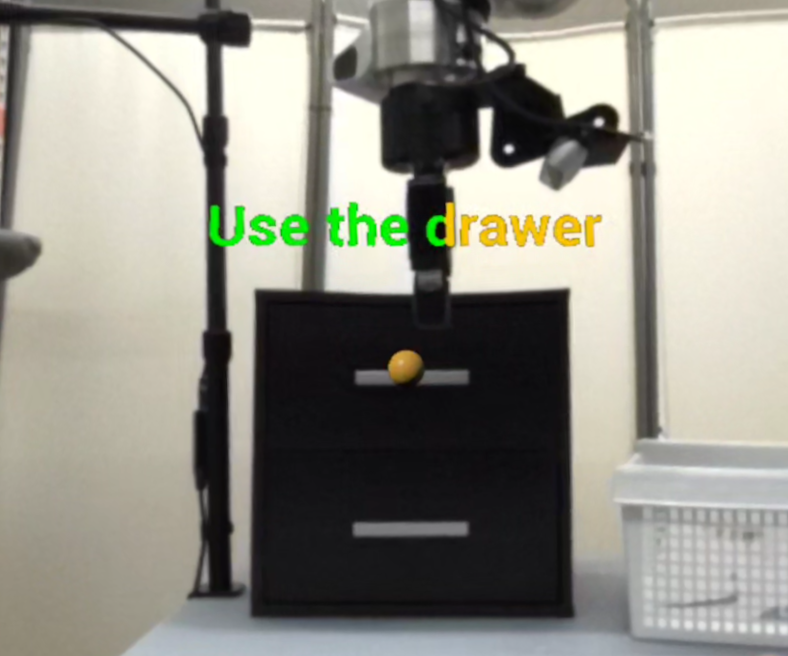}
        \label{fig:confirmation}
    \end{subfigure}
    \caption{Our AR BRI user interface. (Left) The user first selects the object of interest, in this case the drawer, by fixating on it for 3 seconds; their gaze is indicated by the yellow reticule. Once the object is selected, two progress bars appear around the reticule, for the ``Place'' and ``Use'' actions. The bars fill in based on classifications from the MI decoder. (Right) Upon completion of the EEG decoding, the progress bars are hidden and the user is informed of the command that will be sent to the robot using the command text itself. The text acts as a progress bar that fills up in 5 seconds; in case the command displayed is not the intended one, the user can invert the action selection by looking away, leading to the opposite robot action being executed on the selected object.}
    \label{fig:mi_confirmation}
\end{figure}

\subsection{Gaze-Based Object Selection \& Look Away Cancellation}
\label{sec:mrbri:object_selection_cancellation}
The visual interface consists of a user gaze indicator (yellow reticule), object/action feedback text, and an MI interface (Fig. \ref{fig:mi_confirmation}). The virtual overlays (widgets) and text are anchored along the gaze axis to maintain spatial coherence between virtual overlays and detected objects. All the interface components are therefore tangent to a virtual sphere located at the headset's spatial position.

To enable near-real-time object selection, the system continuously monitors the user’s gaze direction through the AR headset’s integrated eye-tracking module at the rendering frequency of the VR headset of \SI{72}{\hertz}.
When the user fixates on a detectable object for the predefined dwell time of three seconds (empirically, we found that this offered a comfortable trade-off between object selection accuracy and responsiveness of the system), the name of the object is displayed above the fixated point. This feedback mechanism was designed to enhance the transparency of the system and reduce uncertainty during object targeting.
Then, the two action options become available as spatial overlays anchored around the selected object, as illustrated in Fig.~\ref{fig:mi_confirmation} (Left).

To allow users to disengage from an unintended object selection, the system supports a cancellation mechanism based on gaze redirection (looking away far enough from the fixation point, toward a corner of the headset's field of view for example). If the user shifts their gaze away from the selected object before the robot action is confirmed (Fig.~\ref{fig:mi_confirmation} (Left)), the MI interface is removed, and the user can proceed to select an object again. This approach provides a lightweight error recovery mechanism without requiring additional control commands.
In case the robot action was already selected (Fig.~\ref{fig:mi_confirmation} (Right)), the look-away cancellation sends the opposite command to the robotics module for execution. This fallback method was used in prior BRI works~\cite{kim2025noir,douglas2025levels} to compensate for potentially poor EEG decoding. Whilst all the user inputs could be implemented solely via eye-tracking, separating control across input modalities distributes the interaction demands on the user, which has been shown to result in improved task performance and reduced perceived workload~\cite{oviatt2004we,triantafyllidis2020study}.

\subsection{Visual Overlays and Object Detection}
\label{sec:mrbri:object_detection}
For users to select objects, the system needs to detect what they are focusing on.
The Meta Quest Pro streams its passthrough view to the object detection manager, which passes it to the object detection module.
This module runs an instance of the OWLv2~\cite{minderer2024scaling} zero-shot object detection model. OWLv2 was chosen as it is an open weights model with zero-shot capability, bypassing the need for task-specific data collection and retraining.

Every three seconds, this modules receives a set of candidate labels corresponding to the objects in the user's physical workspace and the passthrough view.
We filter the resulting detected objects and bounding boxes to return the one (if any) that intersects the user's gaze.
The name of the detected object is then displayed as a textual cue over the reticule.

\subsection{Motor Imagery-Based Action Execution}
\label{sec:mrbri:mi_methods}
Action selection was implemented as an MI-based EEG decoding pipeline.
Two high-level manipulation intents, (``Place'' and ``Use'') were mapped to distinct MI classes (left and right, respectively). After gaze-based object selection, the two action labels were displayed as overlays near the selected object, positioned to the left and right of the gaze reticule. This spatial arrangement was designed to align with the underlying MI paradigm, ensuring consistency between the interface layout and the left/right MI encoding scheme.

To capture EEG signals, we used a g.tec g.GAMMAcap2 cap, with active g.SCARABEO electrodes arranged following the extended international 10–20 system. We kept impedance below \SI{5}{\kilo\ohm} to ensure signal integrity. 22 channels were captured, sampled at a frequency of \SI{256}{\hertz}. The raw signal was preprocessed with a notch filter (\SI{48}-\SI{52}{\hertz}) to remove line noise, followed by the MOABB library's~\cite{jayaram2018moabb} default MI pipeline, which applies bandpass filtering (\SI{8}-\SI{32}{\hertz}), epoching, and signal standardisation. We extracted features via the filter bank common spatial pattern (FBCSP) algorithm~\cite{koles1990spatial,ang2008filter}. A bank of six bandpass filters spanning \SI{8}-\SI{32}{\hertz} was applied to decompose the EEG signals, covering the alpha (\SI{8}-\SI{13}{\hertz}) and beta (\SI{13}-\SI{30}{\hertz}) frequency bands known to be most informative for MI. Four CSP components were then extracted from each frequency band, yielding \SI{24} features per trial in total.
A support vector machine (SVM) with a radial basis function (RBF) kernel and regularization parameter $C$=\SI{0.5} served as the classifier, producing probability estimates for left- and right-hand MI. This decoding pipeline was chosen as it has achieved competitive accuracy on MI benchmarks~\cite{jayaram2018moabb}, while requiring relatively little computational resources.

For each participant, the EEG classifier was trained from a short calibration dataset collected at the beginning of the experimental session, following the protocol used in prior BRI work~\cite{douglas2025levels}. The calibration consisted of 48 trials in total, split into two short sub-sessions, with the 2 classes presented in random order. Each trial displayed a single visual cue---either ``Place'' or ``Use'''---for 4 seconds, providing enough time for participants to perform sustained motor imagery while keeping the calibration phase short enough for practical use. For the ``Place" class, positioned on the left, users had to image a pushing motion away from themselves with their left arm, corresponding to placing an object away. For the ``Use'' class, positioned on the right, users were instructed to imagine a pulling motion toward themselves with their right hand, corresponding to actions such as drinking from a mug or opening a drawer. Each trial was followed by a 1 second ``Rest'' cue before proceeding to the next, with a longer break of 4 seconds every four trials to reduce fatigue and help participants reset their motor imagery.

For online decoding, the continuous LSL stream of EEG signals was processed using the same preprocessing and feature-extraction pipeline that was used for data collection. The online decoding prediction was based on a sliding window approach, where user intent prediction was triggered only if the same class was predicted three times consecutively. This mechanism helped suppress erroneous outputs and improved the reliability of real-time decisions~\cite{douglas2025levels}.

\subsection{Shared Autonomy and Robot Control}

\begin{figure*}[t!]
    \centering
    \includegraphics[width=\linewidth]{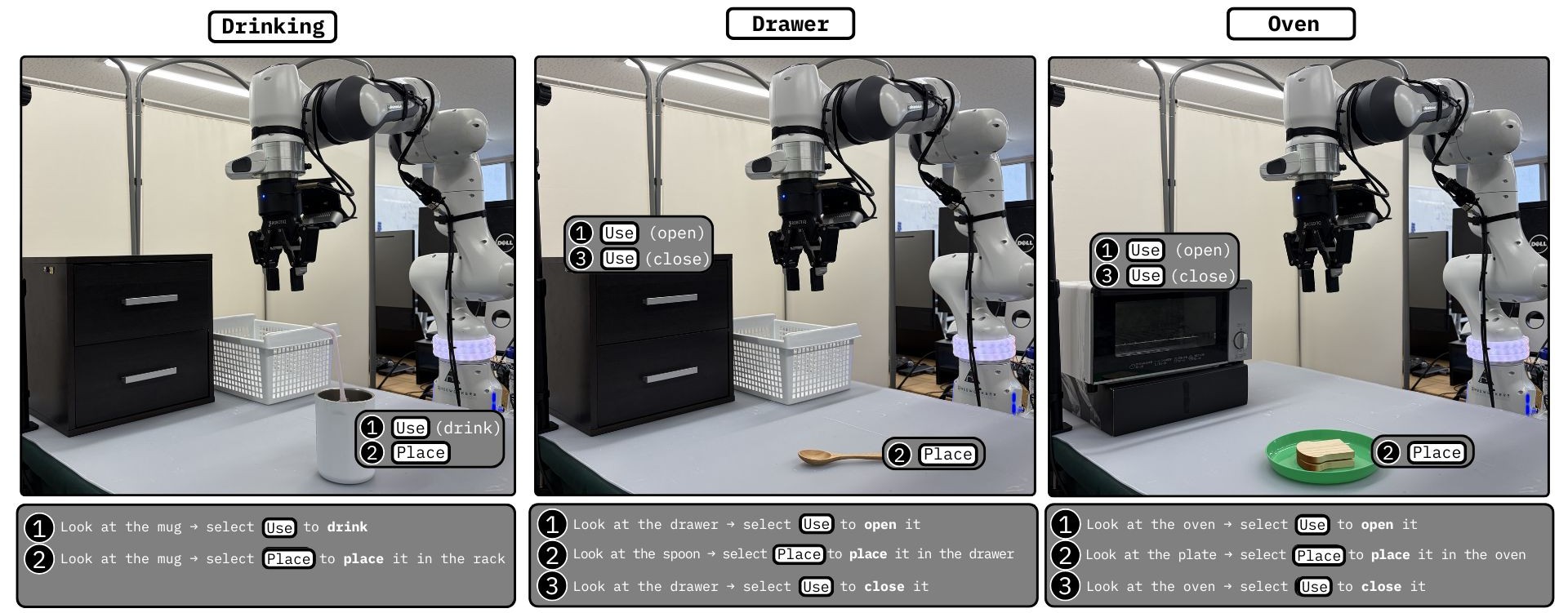}
    \caption{Experimental task setups. The three multi-step manipulation tasks performed in our experiments are illustrated. (Left) \textbf{Drinking Task:} the user drinks from the mug and then places it in the rack. (Center) \textbf{Drawer Task:} the user opens the drawer, places the spoon inside, and closes it. (Right) \textbf{Oven Task:} the user opens the oven, inserts the plate, and closes it. The figure also illustrates the sequence of actions required for each step, indicating the order of object and action (``Place'' or ``Use'') selections for the robot commands to be executed.}
    \label{fig:mr_setup}
\label{fig:setup}
\end{figure*}

To execute the actions selected by the user on a real robot, we use a fine-tuned $\pi_{0.5}$ model~\cite{intelligence2025pi}. We selected $\pi_{0.5}$ for its publicly available weights, strong generalisation across manipulation tasks, and data-efficient fine-tuning.

The input to the model consists of three images (front facing camera, wrist camera and left-back ego camera), and the proprioceptive state of the robot $\left(\bm{x}, \bm{r}, g\right)$, where $\bm{x} \in \mathbb{R}^3$ is the end effector position, $\bm{r} \in \mathbb{R}^3$ is the end effector orientation in axis-angle representation, and $g \in \left[ 0, 1\right]$ is the gripper state.
The action space is $\left(\Delta\bm{x}, \Delta\bm{r}, g_{cmd}\right)$, where $\Delta\bm{x}, \Delta\bm{r}, g_{cmd}$ are deltas to the end effector pose and the gripper command, respectively. We set the model to predict an action chunk size of $25$, with a re-plan horizon of 10, and a control frequency of \SI{12.5}{\hertz}. The actions are executed by a joint-level controller running at \SI{1000}{\Hz} that tracks the targets from the VLA. As this policy is deployed in close proximity to the user, to maintain safety during robot operation we use the operational space control barrier function (OSCBF) framework~\cite{morton2025safe0}.
With a negligible extra computational cost, we constrain the robot’s motion via OSCBF to remain well-behaved close to singularities and within safe bounds to avoid getting too close to the user, maintaining reliable behavior even in critical situations.

Based on prior research~\cite{gu2025safe, yang2025seqvla}, we augmented the $\pi_{0.5}$ model output head to also predict a task progress indicator, $p \in \mathbb{R}$. This allows the robot to reset and be ready to accept new commands autonomously without the need for experimenter intervention, ensuring a smooth flow between subsequent tasks for the users. In practice, task success is detected based on a moving average of the progress returned by the policy. However, as our method does not detect policy failures, we also enforce a timeout-based reset after \SI{40}{\s}.

To fine-tune the model, one teleoperator collected around \SI{90}{\minute} (410 episodes) of in-house data via teleoperation with a gamepad across 9 tasks, 8 of which are used during the user studies, and trained the model for $30,000$ gradient steps with a batch size of $128$. To generate dense labels for task progress prediction, we used values linearly interpolating from 0 to 1 starting at $5\%$ and ending at $95\%$ of each episode's length, filling with zeros and ones respectively at each end of the demonstrations.

\section{EXPERIMENTS}
\label{sec:mrbri:experiment_settings}
To assess the feasibility of our AR BRI system, we conducted a user study with healthy participants performing multi-step manipulation tasks representative of ADLs.
The study aimed to assess (i) whether participants could successfully command actions in sequence using the system (ii) the success rate of the overall system, and (iii) the perceived usability of the interface.

\subsection{Study Procedure and Participants}
\label{sec:mrbri:study_procedure_participants}
A total of 18 healthy participants (6 females, 12 male, mean age = 32.4 ± 7.3 years) participated in the study. Participants had a range of experience with BCIs, from no experience, to having participated in many studies; approximately half considered themselves experienced.
% Participants had a range of experience with brain–computer interfaces, from none to experienced, with half of the participants reporting prior experience at an intermediate or advanced level.

Each experimental session was based on the following standardised protocol.
Upon arrival, the participant received instructions about the study and provided written informed consent. They were then introduced to the MI paradigm and to the structure of the experimental tasks.
Next, the participant was fit with the EEG cap and AR headset, before proceeding to the data collection phase for the EEG-based intent decoder (see Section~\ref{sec:mrbri:mi_methods}). The data collection consisted of two 5 minute sub-sessions, with a 3 minute break in between.

After training and loading the EEG decoder model, participants were given up to 5 minutes of open-ended practice to familiarise themselves with the interface, task flow and required sequence of actions to complete them.
The participants then proceeded to perform the three multi-step tasks in a random order.

After completing the tasks, users completed two questionnaires, and were then debriefed. Our study was approved by the Shiba Palace Clinic Ethics Review Committee.

\subsection{Experimental Setting and Tasks}
\label{sec:expeirment_setting_tasks}
\begin{figure*}[t!]
\centering
    \begin{subfigure}[t]{0.515\linewidth}
        \centering
        \includegraphics[width=\linewidth]{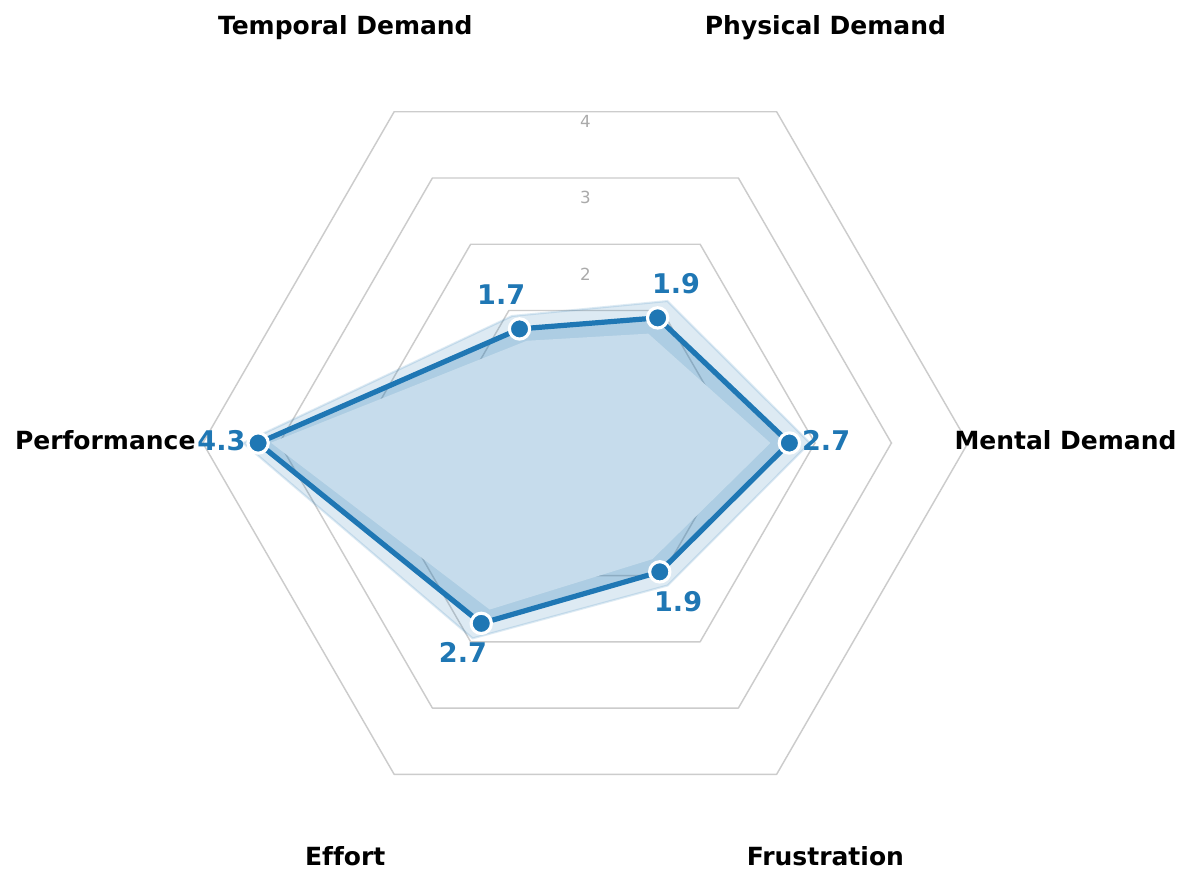}
        \caption{NASA-TLX workload subscale scores.}
        \label{fig:nasa_tlx}
    \end{subfigure}
    \hfill
    \begin{subfigure}[t]{0.475\linewidth}
        \centering
        \includegraphics[width=\linewidth]{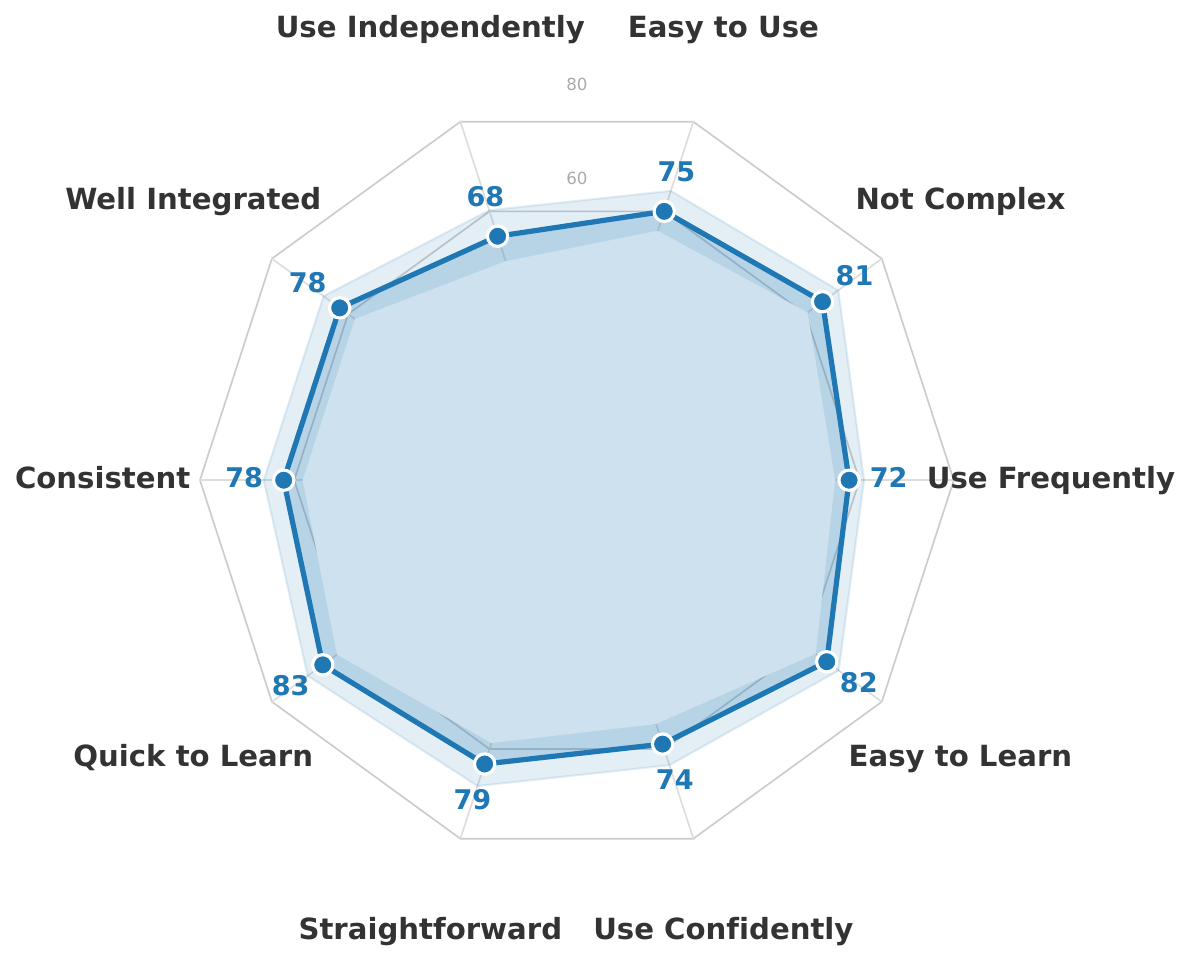}
        \caption{SUS usability subscale scores.}
        \label{fig:sus}
    \end{subfigure}
\caption{(a) NASA-TLX and (b) SUS survey results from the 18 study participants. Mean values are given in bold, with light shading indicating 1 standard deviation. Users reported low workload and high usability for our system.}
\label{fig:questionnaires}
\end{figure*}

Participants were seated next to the robotic arm, facing a table-based workspace. The experiment included two physical settings containing everyday objects. In the first setting, the workspace included a drawer, a dish rack used as a placement container, and either a mug or a spoon placed on the table.
In the second setting, the workspace contained a small oven and a plate with food. Participants performed three different multi-step manipulation tasks (Fig.~\ref{fig:mr_setup}):

\noindent\textbf{(1) Drinking Task:} The participant had to first bring the mug to their mouth to drink, and then place the glass inside the dish rack.

\noindent\textbf{(2) Drawer Task:} The participant had to open the drawer, place the spoon inside it, and then close the drawer.

\noindent\textbf{(3) Oven Task:} The participant had to open the oven, insert the plate with food inside it, and close the oven.

The order of the tasks was randomised across participants. While we defined a max completion time limit of 5 minutes, the longest task completion took at most half that time.

\subsection{Metrics}

We collected different types of metrics, which we present here in a three-category taxonomy \cite{steinfeld2006common}:

\noindent\textbf{Human metrics} included subjective assessments collected through questionnaires administered after the experiment.
Specifically, participants completed the system usability scale (SUS; scored 0--100)~\cite{brooke1996sus} to evaluate the perceived usability, and the raw (unweighted) NASA Task Load Index (NASA-TLX; scored 0--5)~\cite{hart1988development, byers1989traditional} to assess perceived workload.

\noindent\textbf{Robot metrics} captured the performance of the robot during task execution. All subtasks are reported in Table~\ref{tab:subtask_performance}.
For the ``Open'' and ``Close'' subtasks, success is achieved when the container is fully opened or closed, respectively; for the ``Place'' subtasks, the target object must be released in the target receptacle; for the ``Use Mug'' subtask, the robot must bring the mug in proximity of the user's face.

\noindent\textbf{System and interaction metrics} evaluated the overall performance of the human–robot interaction. These included task completion time and task success, as well as the success rate of the EEG decoder.

\section{Results}
\label{sec:results}

\subsection{Robot and System Performance}
\label{sec:results:system_robot_performance}

We report the success rate and execution time for each subtask, as well as overall task completion time, in Table~\ref{tab:subtask_performance}. Both the ``Drawer'' and ``Oven'' tasks were completed with a 100\% success rate. For two users the robot policy failed on the ``Place Mug'' subtask and they had to resend the instruction, resulting in a success rate of $18/20$.

In terms of timing, most policies exhibited low variability, indicating stable and predictable execution. ``Open'' subtasks had the highest variability, reflecting the difficulty of finding a secure grasp for opening. Overall, the ``Drink'' task was the fastest to complete on average, primarily due to having only two subtasks. However, the other two tasks did not take much longer to complete.

Across all three tasks and eight subtasks, the system demonstrated consistent and reliable performance, supporting its viability for real-world assistive manipulation scenarios.

% For all pipelines, model evaluation used a stratified 5-fold cross-validation strategy
With regards to EEG-based intent decoding, offline classification yielded a mean training/validation accuracy of $0.69 \pm 0.16$ and a test accuracy of $0.70 \pm 0.17$, evaluated using a stratified 5-fold cross-validation strategy. 
The post-calibration performance ranged from \SI{60}\% to \SI{94}\% across participants.
This variability reflects well-documented individual differences in MI aptitude~\cite{jeunet2016advances}, suggesting that while some users are able to effectively modulate sensorimotor rhythms, others show limited or no control---a phenomenon often referred to as brain-computer interface illiteracy \cite{becker2022bci}.

Online decoding accuracy improved substantially to $0.86 \pm 0.23$, attributable to two complementary mechanisms: the sliding window scheme applied during the online decoding, which reduces the impact of single erroneous predictions, and the gaze-based error recovery mechanism, which allows users to override incorrect classifications by looking away to cancel the current decoding. Together, these mechanisms ameliorate raw EEG decoding accuracy, partially compensating for inter-session variability in MI performance.

\begin{table}[t]
    \centering
    \caption{Subtask-level success rate and execution time (in seconds), grouped by task
              type, and overall task completion time. We report the mean $\pm$ 1 standard deviation from data across all sessions. The system achieved nearly a 100\% success rate, with the max completion time under 3 minutes.}
    \label{tab:subtask_performance}
    \setlength{\tabcolsep}{2.5pt} 
    \begin{tabular}{llccc}
        \toprule
        \textbf{Task} & \textbf{Subtask} &
        \textbf{Success Rate} & \textbf{Exec.\ Time (s)} &
        \textbf{Comp.\ Time (s)} \\
        \midrule
        Drink  & Use Mug & $1.0$ (18/18) & $18.2 \pm 1.6$ & \multirow{2}{*}{$112.2 \pm 24.6$} \\
               & Place Mug & $0.9$ (18/20) & $20.9 \pm 3.2$ & \\
        \midrule
        Drawer & Open Drawer & $1.0$ (18/18) & $18.8 \pm 4.5$ & \multirow{3}{*}{$131.9 \pm 30.6$} \\
               & Place Spoon & $1.0$ (18/18) & $17.3 \pm 1.4$ & \\
               & Close Drawer & $1.0$ (18/18) & $8.7 \pm 0.8$ & \\
        \midrule
        Oven   & Open Oven & $1.0$ (18/18) & $14.1 \pm 5.2$ & \multirow{3}{*}{$115.9 \pm 15.3$} \\
               & Place Plate & $1.0$ (18/18) & $20.2 \pm 1.4$ & \\
               & Close Oven & $1.0$ (18/18) & $11.3 \pm 0.8$ & \\
        \bottomrule
    \end{tabular}
\end{table}

\subsection{System Usability and Workload Evaluation}
\label{sec:results:nasa_sus}

The NASA-TLX and SUS survey results are shown in Fig.~\ref{fig:questionnaires}. Among all the workload components, the mental demand and effort were the dominant workload contributors, which is expected for a BRI system that requires a sustained cognitive engagement. The physical and temporal demand, however, were relatively low, due to the efficacy of the combined gaze- and MI-driven system. The performance was the highest-rated subscale, suggesting that participants felt they performed well---a positive indicator despite elevated workload perceptions. Finally, the user frustration was also low, suggesting that the system did not cause significant user discomfort despite its cognitive demands. Overall, the workload profile is consistent with a cognitively engaging but physically undemanding assistive system.

The overall SUS score of 76.94 falls in the ``Good'' usability range (70–85), approaching the ``Excellent'' threshold. The subscale scores are notably balanced, with all items being > 67. ``Quick to Learn'' (83) and ``Easy to Learn'' (82) were the highest rated items, indicating an intuitive and easy to learn control system.
The ``Not Complex'' (81) and ``Straightforward'' (79) items follow, suggesting a perceived simplicity of the system, aligning with the ease of learning and usability.
The ``Well Integrated'' (78) and ``Consistent'' (78) items suggest a seamless combination of the various input modalities of the system, coupled with reliable robot control policies, corroborating the robot performance metrics in Table~\ref{tab:subtask_performance}.
The lowest item, ``Use Independently'' (68), suggest a lower confidence of the user being able to use such system by themselves, without assistance from an technician. As the EEG and robot setups currently require domain expertise, this particular outcome is understandable.

Overall, the combined results suggest a favorable reception of the system in this feasibility study. The interface imposed a moderate cognitive load while maintaining good usability and high self-reported performance.

\section{CONCLUSION}
\label{sec:conclusion}
In this work we presented an AR BRI for generalist robot manipulation, combining gaze-based object selection, MI-based action control, and spatially-aware visual feedback within a shared autonomy framework. The system was evaluated in a feasibility study with 18 healthy participants across three multi-step ADL-inspired tasks. Results demonstrate strong task performance, with near-perfect subtask success rates and highly consistent execution times across all conditions. The SUS score of 76.94 indicates good usability, and participants found the system quick and easy to learn despite the cognitive demands inherent to EEG-based control.
NASA-TLX results confirm a moderate cognitive load profile, while physical and temporal demands remained low, consistent with the passive, gaze- and MI-driven nature of the interface.
Beyond quantitative measures, participants responded positively to the overall experience, expressing genuine engagement with the system owing to the combination of real-world object interaction and robot execution that contributed to an increased sense of immersion.

Nevertheless, several limitations must be acknowledged. On the hardware side, the concurrent use of a separate EEG cap and AR headset introduced wearability challenges, with some participants reporting discomfort from wearing both devices simultaneously.
Future iterations would benefit from more tightly integrated hardware solutions, such as Galea\footnote{https://galea.co/} or Kaptics\footnote{https://www.kaptics.com/} headsets that natively integrate EEG sensors and an AR headset in a single, ergonomic form factor.
More critically, all participants in this study were healthy adults, and none belonged to the target demographic of individuals with motor impairments. While the results establish a promising proof of concept, the system's real-world utility for assistive applications remains to be validated with the intended user population, whose needs, capabilities, and tolerance for cognitive load may differ substantially.

\section*{ACKNOWLEDGMENT}
The authors would like to thank Hannah Kodama Douglas for her assistance during the user study.

\bibliographystyle{IEEEtran}
\bibliography{references.bib}

\end{document}